\documentclass[runningheads]{llncs}
\usepackage{graphicx}
\usepackage{comment}
\usepackage{amsmath,amssymb}
\usepackage{url,multirow,tabulary}
\usepackage{color}
\usepackage{subcaption}
\usepackage[cmintegrals]{newtxmath}
\usepackage{multirow}
\usepackage{tikz}
\usepackage{makecell}
\usepackage{tabulary}
\usepackage{booktabs}
\usepackage{array}
\usepackage{url}

\newcommand{\etal}{\textit{et al.}}
\usepackage[colorlinks,citecolor=black,bookmarks=false]{hyperref}
\usepackage[width=122mm,left=12mm,paperwidth=146mm,height=193mm,top=12mm,paperheight=217mm]{geometry}
\newcommand{\greenline}{\raisebox{2pt}{\tikz{\draw[-,green,solid,line width = 0.9pt](0,0) -- (5mm,0);}}}
\newcommand{\redline}{\raisebox{2pt}{\tikz{\draw[-,red,solid,line width = 0.9pt](0,0) -- (5mm,0);}}}
\newcommand{\blueline}{\raisebox{2pt}{\tikz{\draw[-,blue,solid,line width = 0.9pt](0,0) -- (5mm,0);}}}

\newcolumntype{L}[1]{>{\raggedright\arraybackslash}p{#1}}
\newcolumntype{C}[1]{>{\centering\arraybackslash}p{#1}}
\newcolumntype{R}[1]{>{\raggedleft\arraybackslash}p{#1}}

\begin{document}
\pagestyle{headings}
\mainmatter

\title{SNE-RoadSeg: Incorporating Surface Normal Information into Semantic Segmentation for Accurate Freespace Detection}

\titlerunning{SNE-RoadSeg}
\authorrunning{R. Fan, H. Wang, P. Cai, and M. Liu}

\author{Rui Fan\inst{1}$^\star$\orcidID{0000-0003-2593-6596} \and
    Hengli Wang\inst{2}\thanks{These authors contributed equally to this work and are therefore joint first authors.}\orcidID{0000-0002-7515-9759} \and\\
    Peide Cai\inst{2}\orcidID{0000-0002-9759-2991} \and Ming Liu\inst{2}\orcidID{0000-0002-4500-238X}}
\institute{UC San Diego \\
    \email{rui.fan@ieee.org} \and
    HKUST Robotics Institute \\
    \email{\{hwangdf, peide.cai, eelium\}@ust.hk}}

\maketitle

\begin{abstract}
Freespace detection is an essential component of visual perception for self-driving cars. The recent efforts made in data-fusion convolutional neural networks (CNNs) have significantly improved semantic driving scene segmentation. Freespace can be hypothesized as a ground plane, on which the points have similar surface normals. Hence, in this paper, we first introduce a novel module, named surface normal estimator (SNE), which can infer surface normal information from dense depth/disparity images with high accuracy and efficiency. Furthermore, we propose a data-fusion CNN architecture, referred to as RoadSeg, which can extract and fuse features from both RGB images and the inferred surface normal information for accurate freespace detection. For research purposes, we publish a large-scale synthetic freespace detection dataset, named Ready-to-Drive (R2D) road dataset, collected under different illumination and weather conditions. The experimental results demonstrate that our proposed SNE module can benefit all the state-of-the-art CNNs for freespace detection, and our SNE-RoadSeg achieves the best overall performance among different datasets.
\end{abstract}

\begin{center}
\textbf{Source Code, Dataset and Demo Video:}\\
\url{sites.google.com/view/sne-roadseg}
\end{center}

\section{Introduction}
\label{sec.introduction}
Autonomous cars are a regular feature in science fiction films and series, but thanks to the rise of artificial intelligence, the fantasy of picking up one such vehicle  at your garage forecourt has turned into reality. Driving scene understanding is a crucial task for autonomous cars, and it has taken a big leap with recent advances in artificial intelligence \cite{alvarez2012road}. Collision-free space (or simply freespace) detection is a fundamental component of driving scene understanding \cite{sless2019road}. Freespace detection approaches generally classify each pixel in an RGB or depth/disparity image as drivable or undrivable. Such pixel-level classification results are then utilized by other modules in the autonomous system, such as trajectory prediction \cite{cai2020prob} and path planning \cite{thoma2019mapping}, to ensure that the autonomous car can navigate safely in complex environments.

The existing freespace detection approaches can be categorized as either traditional or machine/deep learning-based. The traditional approaches generally formulate freespace with an explicit geometry model and find its best coefficients using optimization approaches \cite{fan2019pothole}. \cite{wedel2009b} is a typical traditional freespace detection algorithm, where road segmentation is performed by fitting a B-spline model to the road disparity projections on a 2D disparity histogram (generally known as a v-disparity image) \cite{fan2019uav}. With recent advances in machine/deep learning, freespace detection is typically regarded as a semantic driving scene segmentation problem, where the convolutional neural networks (CNNs) are used to learn its best solution \cite{wang2020applying}. For instance, Lu {\etal} \cite{lu2019monocular} employed an encoder-decoder architecture to segment RGB images in the bird's eye view for end-to-end freespace detection. Recently, many researchers have resorted to data-fusion CNN architectures to further improve the accuracy of semantic image segmentation. For example, Hazirbas {\etal} \cite{hazirbas2016fusenet} incorporated depth information into conventional semantic segmentation via a data-fusion CNN architecture, which greatly enhanced the performance of driving scene segmentation.

In this paper, we first introduce a novel module named surface normal estimator (SNE), which can infer surface normal information from dense disparity/depth images with both high precision and efficiency. Additionally, we design a data-fusion CNN architecture named RoadSeg, which is capable of incorporating both RGB and surface normal information into semantic segmentation for accurate freespace detection. Since the existing freespace detection datasets with diverse illumination and weather conditions do not have either disparity/depth information or freespace ground truth, we created a large-scale synthetic freespace detection dataset, named Ready-to-Drive (R2D) road dataset (containing 11430 pairs of RGB and depth images), under different illumination and weather conditions. Our R2D road dataset is also publicly available for research purposes.  To validate the feasibility and effectiveness of our introduced SNE module, we use three road datasets (KITTI \cite{Fritsch2013ITSC}, SYNTHIA \cite{HernandezBMVC17} and our R2D) to train ten state-of-the-art CNNs (six single-modal CNNs and four data-fusion CNNs), with and without our proposed SNE module embedded. The experiments demonstrate that our proposed SNE module can benefit all these CNNs for freespace detection. Also, our method SNE-RoadSeg outperforms all other CNNs for freespace detection, where its overall performance is the {second} best on the KITTI road benchmark\footnote{\url{cvlibs.net/datasets/kitti/eval_road.php}} \cite{Fritsch2013ITSC}.

The remainder of this paper is structured as follows: Section \ref{sec.related_work} provides an overview of the state-of-the-art CNNs for semantic image segmentation. Section \ref{sec.sne-roadseg} introduces our proposed SNE-RoadSeg. Section \ref{sec.experiments} shows the experimental results and discusses both the effectiveness of our proposed SNE module and the performance of our SNE-RoadSeg. Finally, Section \ref{sec.conclusion} concludes the paper.

\section{Related Work}
\label{sec.related_work}
In 2015, Long {\etal} \cite{long2015fully} introduced Fully Convolutional Network (FCN), a CNN for end-to-end semantic image segmentation. Since then, research on this topic has exploded. Based on  FCN, Ronneberger {\etal} \cite{ronneberger2015u} proposed U-Net in the same year, which consists of a contracting path and an expansive path \cite{ronneberger2015u}. It adds skip connections between the contracting path and the expansive path to help better recover the full spatial resolution. Different from  FCN, SegNet \cite{badrinarayanan2017segnet} utilizes an encoder-decoder architecture, which has become the mainstream structure for following approaches. An encoder-decoder architecture is typically composed of an encoder, a decoder and a final pixel-wise classification layer.

Furthermore, DeepLabv3+ \cite{chen2018encoder}, developed from DeepLabv1 \cite{chen2014semantic}, DeepLabv2 \cite{chen2017deeplab} and DeepLabv3 \cite{chen2017rethinking}, was proposed in 2018. It employs depth-wise separable convolution in both atrous spatial pyramid pooling (ASPP) and the decoder, which makes its encoder-decoder architecture much faster and stronger \cite{chen2018encoder}. Although the ASPP can generate feature maps by concatenating multiple atrous-convolved features, the resolution of the generated feature maps is not sufficiently dense for some applications such as autonomous driving \cite{chen2017deeplab}. To address this problem, DenseASPP \cite{yang2018denseaspp} was designed to connect atrous convolutional layers (ACLs) densely. It is capable of generating multi-scale features that cover a larger and denser scale range, without significantly increasing the model size \cite{yang2018denseaspp}.

Different from the above-mentioned CNNs, DUpsampling \cite{tian2019decoders} was proposed to recover the pixel-wise prediction by employing a data-dependent decoder. It allows the decoder to downsample the fused features before merging them, which not only reduces computational costs, but also decouples the resolutions of both the fused features and the final prediction \cite{tian2019decoders}. GSCNN \cite{takikawa2019gated} utilizes a novel two-branch architecture consisting of a regular (classical) branch and a shape branch. The regular branch can be any backbone architecture, while the shape branch processes the shape information in parallel with the regular branch. Experimental results have demonstrated that this architecture can significantly boost the performance on thinner and smaller objects \cite{takikawa2019gated}.

FuseNet \cite{hazirbas2016fusenet} was designed to use RGB-D data for semantic image segmentation. The key ingredient of FuseNet is a fusion block, which employs element-wise summation to combine the feature maps obtained from two encoders. Although FuseNet \cite{hazirbas2016fusenet} demonstrates impressive performance, the ability of CNNs to handle geometric information is limited, due to the fixed grid kernel structure \cite{wang2018depth}. To address this problem, depth-aware CNN \cite{wang2018depth} presents two intuitive and flexible operations: depth-aware convolution and depth-aware average pooling. These operations can efficiently incorporate geometric information into the CNN by leveraging the depth similarity between pixels \cite{wang2018depth}.

MFNet \cite{ha2017mfnet} was proposed for semantic driving scene segmentation with the use of RGB-thermal vision data. In order to meet the real-time requirement of autonomous driving applications, MFNet focuses on minimizing the trade-off between accuracy and efficiency. Similarly, RTFNet \cite{sun2019rtfnet} was developed to improve the semantic image segmentation performance using RGB-thermal vision data. Its main contribution is a novel decoder, which leverages short-cuts to produce sharp boundaries while keeping more detailed information \cite{sun2019rtfnet}.

\section{SNE-RoadSeg}
\label{sec.sne-roadseg}

\begin{figure*}[t]
    \centering
    \includegraphics[width=\textwidth]{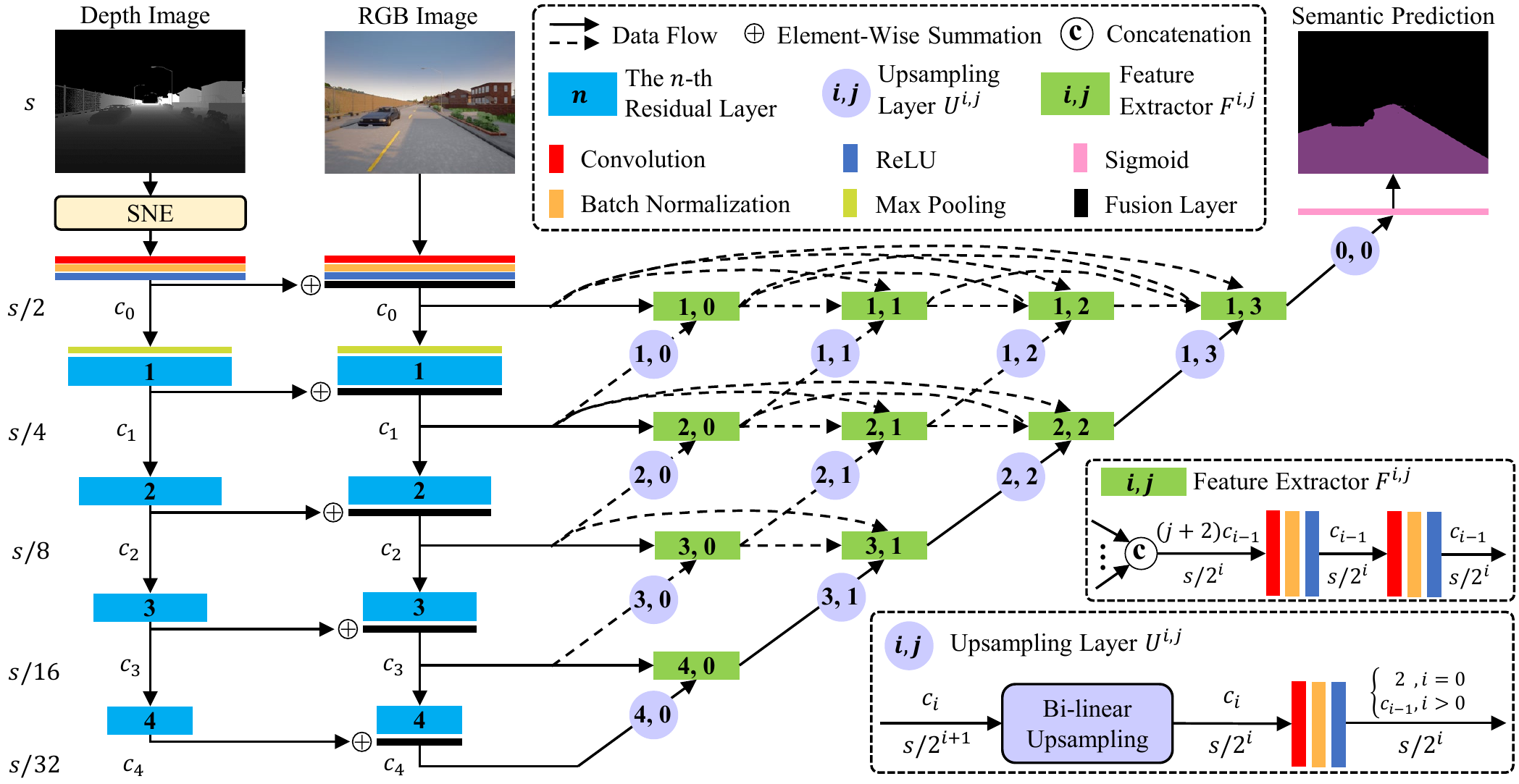}
    \caption{The architecture of our SNE-RoadSeg. It consists of our SNE module, an RGB encoder, a surface normal encoder and a decoder with densely-connected skip connections. $s$ represents the input resolution of the RGB and depth images. $c_n$ represents the number of feature map channels at different levels.}
    \label{fig.roadseg}
\end{figure*}

\subsection{SNE}
\label{sec.sne}
The proposed SNE is developed from our recent work three-filters-to-normal (3F2N) \cite{fan2020three}.
Its architecture is shown in Fig. \ref{fig.sne}. For a perspective camera model, a 3D point $\mathbf{P}=[X,Y,Z]^\top$ in the Euclidean coordinate system can be linked with a 2D image pixel $\mathbf{p}=[x,y]^\top$ using:
\begin{equation}
Z\begin{bmatrix}
\mathbf{p}\\1
\end{bmatrix}=\mathbf{K}\mathbf{P}=
\begin{bmatrix}
f_x & 0 & x_\text{o}\\
0 & f_y & y_\text{o}\\
0 & 0 & 1
\end{bmatrix}\mathbf{P},
\label{eq.intrinisc_matrix}
\end{equation}
where $\mathbf{K}$ is the camera intrinsic matrix;  $\mathbf{p}_\text{o}=[x_\text{o},y_\text{o}]^\top$ is the image center; $f_x$ and $f_y$ are the camera focal lengths in pixels. The simplest way to estimate the surface normal $\mathbf{n}=[n_x, n_y, n_z]^\top$ of $\mathbf{P}$ is to fit a local plane:
\begin{equation}
n_x X + n_y Y + n_z Z + d = 0
\label{eq.plane_function}
\end{equation}
to $\mathbf{N}_\mathbf{P}^+=[\mathbf{P}, \mathbf{N}_\mathbf{P}]^\top$, where $\mathbf{N}_\mathbf{P}=[\mathbf{Q}_1, \dots, \mathbf{Q}_k]^\top$ is a set of $k$ neighboring points of $\mathbf{P}$. Combining (\ref{eq.intrinisc_matrix}) and (\ref{eq.plane_function}) results in \cite{fan2020three}:
\begin{figure*}[t]
    \centering
    \includegraphics[width=0.98\textwidth]{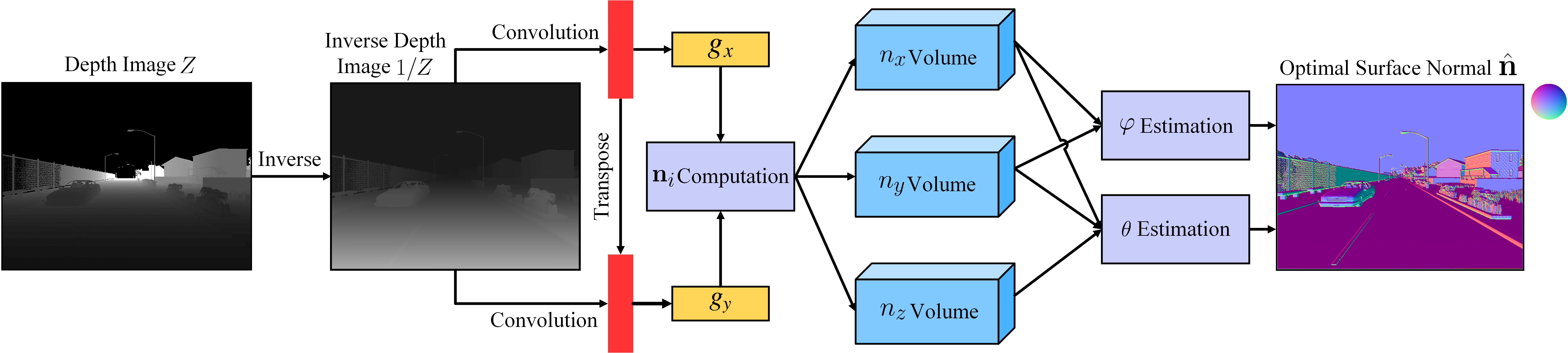}
    \caption{The architecture of our proposed SNE module.}
    \label{fig.sne}
\end{figure*}
\begin{equation}
\frac{1}{Z}=-\frac{1}{d}\bigg(n_x\frac{x-x_\text{o}}{f_x}+n_y\frac{y-y_\text{o}}{f_y}+n_z\bigg).
\label{eq.1/z_n}
\end{equation}
Differentiating (\ref{eq.1/z_n}) with respect to $x$ and $y$ leads to:
\begin{equation}
g_x=\frac{\partial 1/Z}{\partial x}=-\frac{n_x}{d f_x},\ \ \ g_y=\frac{\partial 1/Z}{\partial y}=-\frac{n_y}{d f_y},
\label{eq.d1/z_to_du_dv}
\end{equation}
which, as illustrated in Fig. \ref{fig.sne}, can be respectively  approximated  by convolving the inverse depth image $1/Z$ (or a disparity image, as disparity is in inverse proportion to depth) with a horizontal and a vertical image gradient filter \cite{fan2020three}.
Rearranging (\ref{eq.d1/z_to_du_dv}) results in the expressions of $n_x$ and $n_y$ as follows:
\begin{equation}
n_x=-d f_x g_x, \ \ \ n_y=-d f_y g_y.
\label{eq.nx1_ny1}
\end{equation}
Given an arbitrary $\mathbf{Q}_{i}\in \mathbf{N}_\mathbf{P}$, we can compute its corresponding ${n}_{z_i}$ by plugging (\ref{eq.nx1_ny1}) into (\ref{eq.plane_function}):
\begin{equation}
{n}_{z_i}=d\frac{ f_x \Delta {X_i} g_x + f_y \Delta {Y_i} g_y }{\Delta {Z_i}},
\label{eq.nz1}
\end{equation}
where ${\mathbf{Q}_i}-\mathbf{P}=[\Delta {X_i}, \Delta {Y_i}, \Delta {Z_i}]^\top$.
Since (\ref{eq.nx1_ny1}) and (\ref{eq.nz1}) have a common factor of $-d$, the surface normal $\mathbf{n}_i$ obtained from ${\mathbf{Q}_i}$ and ${\mathbf{P}}$ has the following expression \cite{wang2020applying}:
\begin{equation}
\mathbf{n}_i = \Big[f_x  g_x,\ \ f_y  g_y,\ \ -\frac{ f_x \Delta {X_i} g_x + f_y \Delta {Y_i} g_y }{\Delta {Z_i}} \Big]^\top.
\label{eq.nx2_ny2_nz2}
\end{equation}
A $k$-connected neighborhood system $\mathbf{N}_\mathbf{P}$ of $\mathbf{P}$ can produce $k$ normalized surface normals $\bar{\mathbf{n}}_{1}$, \dots, $\bar{\mathbf{n}}_{k}$, where $\bar{\mathbf{n}}_i=\frac{\mathbf{n}_i}{\|\mathbf{n}_i\|_2}=[\bar{n}_{x_i},\bar{n}_{y_i},\bar{n}_{z_i}]^\top$. Since any normalized surface normals are projected on a sphere with center $(0,0,0)$ and radius $1$, we believe that the optimal surface normal $\hat{\mathbf{n}}$ for $\mathbf{P}$ is also projected somewhere on the same sphere, where the projections of $\bar{\mathbf{n}}_{1}$, \dots, $\bar{\mathbf{n}}_{k}$ distribute most intensively \cite{fan2019pothole}. $\hat{\mathbf{n}}$ can be written in spherical coordinates as follows:
\begin{equation}
\hat{\mathbf{n}} =
\Big[\sin\theta\cos\varphi,\
\sin\theta\sin\varphi,\
\cos\theta\Big]^\top,
\label{eq.spherical_coordinates}
\end{equation}
where $\theta\in[0,\pi]$ denotes inclination and $\varphi\in[0,2\pi)$ denotes azimuth. $\varphi$ can be computed using:
\begin{equation}
\varphi=\arctan\bigg(\frac{f_yg_y}{f_xg_x}\bigg).
\end{equation}
Similar to \cite{fan2019pothole}, we hypothesize that the angle between an arbitrary pair of normalized surface normals is less than $\pi/2$.
$\hat{\mathbf{n}}$ can therefore be estimated by minimizing $E= -\sum_{i=1}^{k}\hat{\mathbf{n}}\cdot{\bar{\mathbf{{n}}}}_{i}$ \cite{fan2019pothole}. $\frac{\partial E}{\partial \theta}=0$ obtains:
\begin{equation}
\theta= \arctan\Bigg(\frac{\sum_{i=1}^{k}\bar{n}_{x_i}\cos\varphi+\sum_{i=1}^{k}\bar{n}_{y_i}\sin\varphi}{\sum_{i=1}^{k}\bar{n}_{z_i}}\Bigg).
\label{eq.theta}
\end{equation}
Substituting $\theta$ and $\varphi$ into (\ref{eq.spherical_coordinates}) results in the optimal surface normal $\hat{\mathbf{n}}$, as shown in Fig. \ref{fig.sne}. The performance of our proposed SNE will be discussed in Section \ref{sec.experiments}.

\subsection{RoadSeg}
\label{sec.cnn}

U-Net \cite{ronneberger2015u} has demonstrated the effectiveness of using skip connections in recovering the full spatial resolution. However, its skip connections force aggregations only at the same-scale feature maps of the encoder and decoder, which, we believe, is an unnecessary constraint. Inspired by DenseNet \cite{huang2017densely}, we propose RoadSeg, which exploits densely-connected skip connections to realize flexible feature fusion in the decoder.

As shown in Fig. \ref{fig.roadseg}, our proposed RoadSeg also adopts the popular encoder-decoder architecture. An RGB encoder and a surface normal encoder is employed to extract the feature maps from RGB images and from the inferred surface normal information, respectively. The extracted RGB and surface normal feature maps are hierarchically fused through element-wise summations. The fused feature maps are then fused again in the decoder through densely-connected skip connections to restore the resolution of the feature maps. At the end of RoadSeg, a sigmoid layer is used to generate the probability map for the semantic driving scene segmentation.

We use ResNet \cite{he2016deep} as the backbone of our RGB and surface normal encoders, the structures of which are identical to each other. Specifically, the initial block consists of a convolutional layer, a batch normalization layer and a ReLU activation layer. Then, a max pooling layer and four residual layers are sequentially employed to gradually reduce the resolution as well as increase the number of feature map channels. ResNet has five architectures: ResNet-18, ResNet-34, ResNet-50, ResNet-101 and ResNet-152. Our RoadSeg follows the same naming rule of ResNet. $c_n$, the number of feature map channels (see Fig. \ref{fig.roadseg}) varies with respect to the adopted ResNet architecture. Specifically, $c_0$--$c_4$ are 64, 64, 128, 256 and 512, respectively, for ResNet-18 and ResNet-34, and are 64, 256, 512, 1024 and 2048, respectively, for ResNet-50, ResNet-101 and ResNet-152.

The decoder consists of two different types of modules: a) feature extractors $F^{i, j}$ and b) upsampling layers $U^{i, j}$, which are connected densely to realize flexible feature fusion. The feature extractor is employed to extract features from the fused feature maps, and it ensures that the feature map resolution is unchanged. The upsampling layer is employed to increase the resolution and decrease the feature map channels. Three convolutional layers in the feature extractor and the upsampling layer have the same kernel size of $3 \times 3$, the same stride of $1$ and the same padding of $1$.

\section{Experiments}
\label{sec.experiments}

\subsection{Datasets and Experimental Setup}
\label{sec.dataset}

In our experiments, we first evaluate the performance of our proposed SNE on the DIODE dataset \cite{vasiljevic2019diode}, a public surface normal estimation dataset containing RGBD vision data of both indoor and outdoor scenarios.  We utilize the average angular error (AAE), $e_\text{AAE}=\frac{1}{m} \sum_{k=1}^{m} \cos ^{-1}\left(\frac{\left\langle\mathbf{n}_{k}, \hat{\mathbf{n}}_{k}\right\rangle}{\left\|\mathbf{n}_{k}\right\|_{2}\left\|\hat{\mathbf{n}}_{k}\right\|_{2}}\right)$,
 to quantify our SNE's accuracy, where $m$ is the number of 3D points used for evaluation; $\mathbf{n}_{k}$ and $\hat{\mathbf{n}}_{k}$ is the ground truth and estimated (optimal) surface normal, respectively. The experimental results are presented in Section \ref{sec.performance_evaluation_of_sne}.

Then, we carry out the experiments on the following three datasets to evaluate the performance of our proposed SNE-RoadSeg for freespace detection:
\begin{itemize}
    \item The KITTI road dataset \cite{Fritsch2013ITSC}: this dataset provides real-world RGB-D vision data. We split it into three subsets: a) training (173 images), b) validation (58 images), and c) testing (58 images).
    \item The SYNTHIA road dataset \cite{HernandezBMVC17}: this dataset provides synthetic RGB-D vision data. We select 2224 images from it and group them into: a) training (1334 images), b) validation (445 images), and c) testing (445 images).
    \item Our R2D road dataset: along with our proposed SNE-RoadSeg, we also publish a large-scale synthetic freespace detection dataset, named R2D road dataset. This dataset is created using the CARLA\footnote{\url{carla.org}} simulator \cite{dosovitskiy2017carla}. Firstly, we mount a simulated stereo rig (baseline: 1.5 m) on the top of a vehicle to capture synchronized stereo images (resolution:  640$\times$480 pixels) at 10 fps. The vehicle navigates in six different scenarios under different illumination and weather conditions (sunny, rainy, day and sunset). There are a total of 11430 pairs of stereo images with corresponding depth images and semantic segmentation ground truth. We split them into three subsets: a) training (6117 images), b) validation (2624 images), and c) testing (2689 images). Our dataset is publicly available at \url{sites.google.com/view/sne-roadseg} for research purposes.
\end{itemize}

We use these three datasets to train ten state-of-the-art CNNs, including six single-modal CNNs and four data-fusion CNNs. We conduct the experiments of single-modal CNNs with three setups: a) training with RGB images, b) training with depth images, and c) training with surface normal images (generated from depth images using our SNE), which are denoted as \textbf{RGB}, \textbf{Depth} and \textbf{SNE-Depth}, respectively. Similarly, the experiments of data-fusion CNNs are conducted using two setups: training using RGB-D vision data, with and without our SNE embedded, which are denoted as \textbf{RGBD} and \textbf{SNE-RGBD}, respectively. To compare the performances between our proposed RoadSeg and other state-of-the-art CNNs, we train our RoadSeg with the same setups as for the data-fusion CNNs on the three datasets. Moreover, we re-train our SNE-RoadSeg for the result submission to the KITTI road benchmark \cite{Fritsch2013ITSC}. The experimental results are presented in Section \ref{sec.performance_evaluation_of_roadseg}. Additionally, the ablation study of our SNE-RoadSeg is provided in Section \ref{sec.ablation_study}.

Five common metrics are used for the performance evaluation of freespace detection: accuracy, precision, recall, F-score and the intersection over union (IoU). Their corresponding definitions are  as follows: Accuracy $=\frac{n_{\mathrm{tp}}+n_{\mathrm{tn}}}{n_{\mathrm{tp}}+n_{\mathrm{tn}}+n_{\mathrm{fp}}+n_{\mathrm{fn}}}$, Precision $=\frac{n_{\mathrm{tp}}}{n_{\mathrm{tp}}+n_{\mathrm{fp}}}$, Recall $=\frac{n_{\mathrm{tp}}}{n_{\mathrm{tp}}+n_{\mathrm{fn}}}$, F-score $=\frac{2 n_{\mathrm{tp}}^{2}}{2 n_{\mathrm{tp}}^{2}+n_{\mathrm{tp}}\left(n_{\mathrm{fp}}+n_{\mathrm{fn}}\right)}$ and IoU $=\frac{n_{\mathrm{tp}}}{n_{\mathrm{tp}}+n_{\mathrm{fp}}+n_{\mathrm{fn}}}$, where $n_{\mathrm{tp}}$, $n_{\mathrm{tn}}$, $n_{\mathrm{fp}}$ and $n_{\mathrm{fn}}$ represents the true positive, true negative, false positive, and false negative pixel numbers, respectively.
In addition, the stochastic gradient descent with momentum (SGDM) optimizer is utilized to minimize the loss function, and the initial learning rate is set to  $0.001$. Furthermore, we adopt the early stopping mechanism on the validation subset to avoid over-fitting. The performance is then quantified using the testing subset.

\begin{figure*}[!t]
    \centering
    \includegraphics[width=\textwidth]{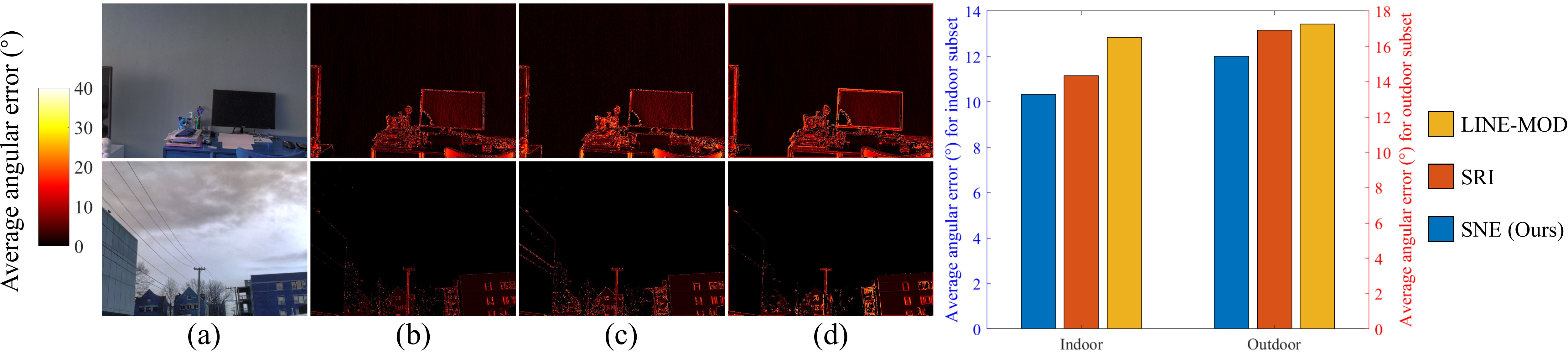}
    \caption{Qualitative and quantitative results on the DIODE dataset: (a) RGB images; (b)--(d): the angular error maps obtained using our proposed SNE, SRI \cite{badino2011fast} and LINE-MOD \cite{hinterstoisser2011gradient}, respectively.}
    \label{fig.diode}
\end{figure*}

\begin{figure*}[!t]
    \centering
    \begin{subfigure}{0.96\textwidth}
        \includegraphics[width=\textwidth]{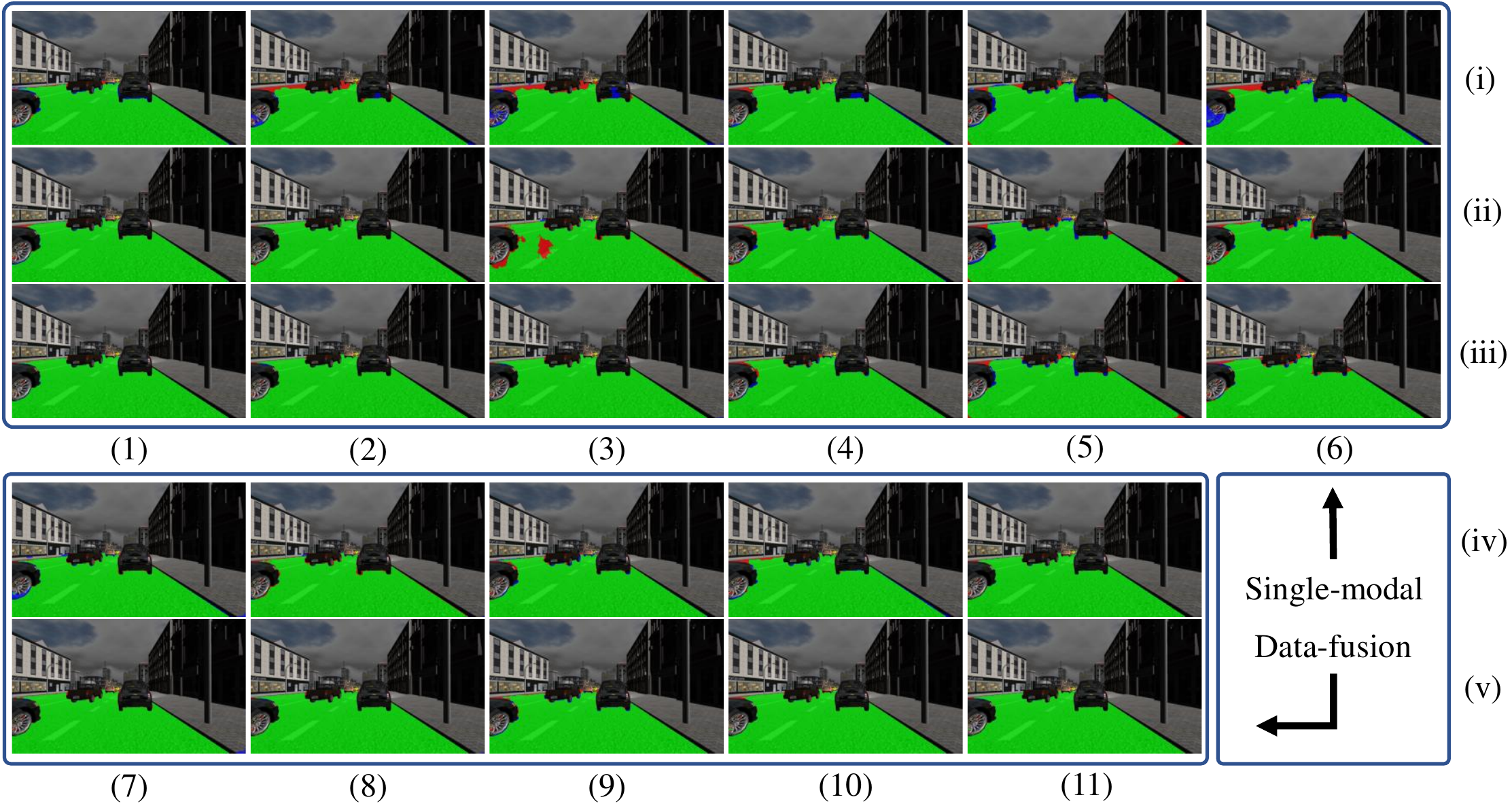}
        \caption{}
        \label{fig.synthia}
    \end{subfigure}
    \begin{subfigure}{0.96\textwidth}
        \includegraphics[width=\textwidth]{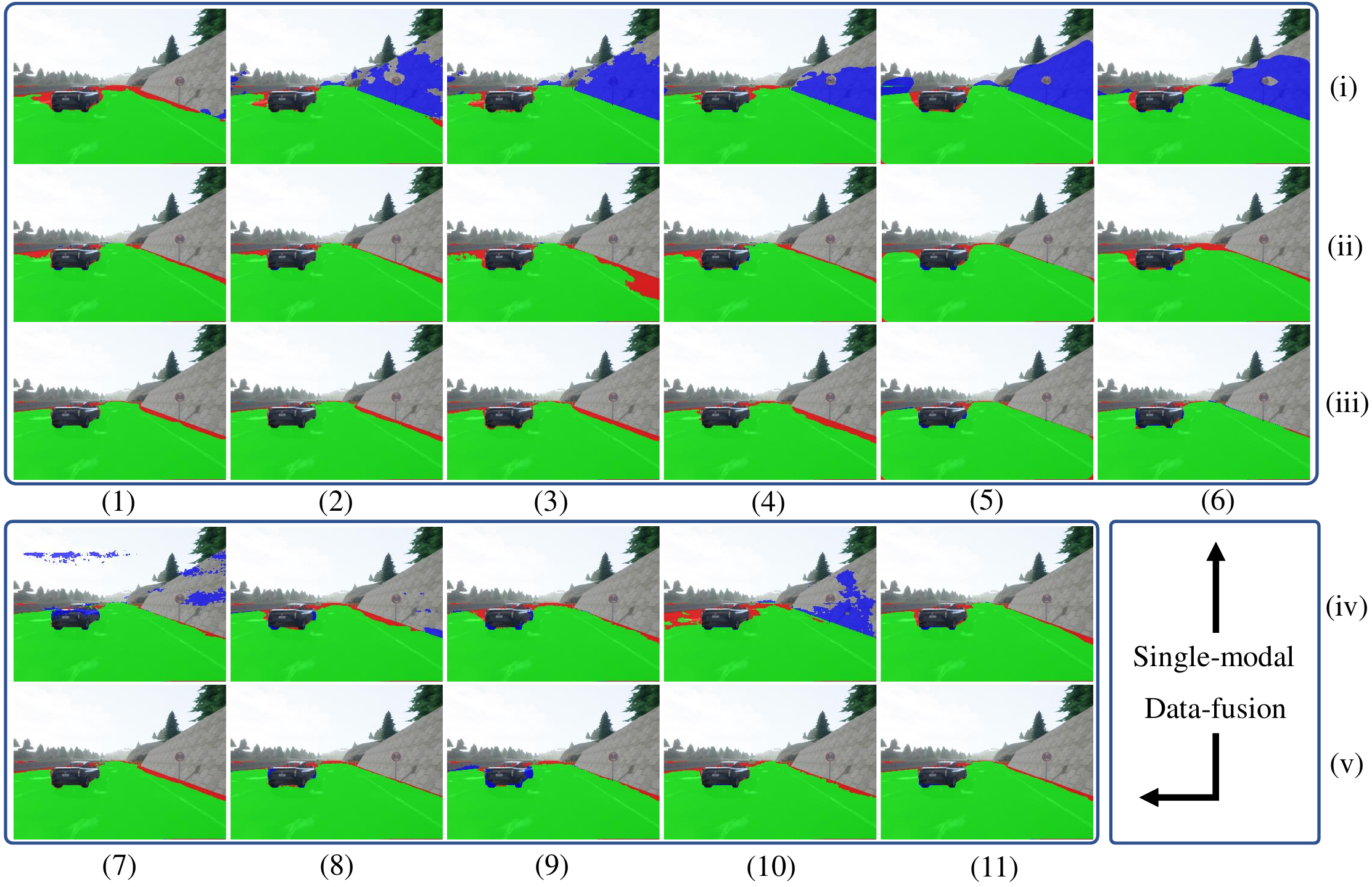}
        \caption{}
        \label{fig.carla}
    \end{subfigure}
    \caption{Examples of the experimental results on (a) the SYNTHIA road dataset and (b) our R2D road dataset: (i) RGB, (ii) Depth, (iii) SNE-Depth (Ours), (iv) RGBD and (v) SNE-RGBD (Ours); (1) DeepLabv3+ \cite{chen2018encoder}, (2) U-Net \cite{ronneberger2015u}, (3) SegNet \cite{badrinarayanan2017segnet}, (4) GSCNN \cite{takikawa2019gated}, (5) DUpsampling \cite{tian2019decoders}, (6) DenseASPP \cite{yang2018denseaspp}, (7) FuseNet \cite{hazirbas2016fusenet}, (8) RTFNet \cite{sun2019rtfnet}, (9) Depth-aware CNN \cite{wang2018depth}, (10) MFNet \cite{ha2017mfnet} and (11) RoadSeg (Ours). The true positive, false negative and false positive pixels are shown in green, red and blue, respectively.}
    \label{fig.synthiapluscarla}
\end{figure*}

\begin{figure*}[!t]
    \centering
    \includegraphics[width=0.86\textwidth]{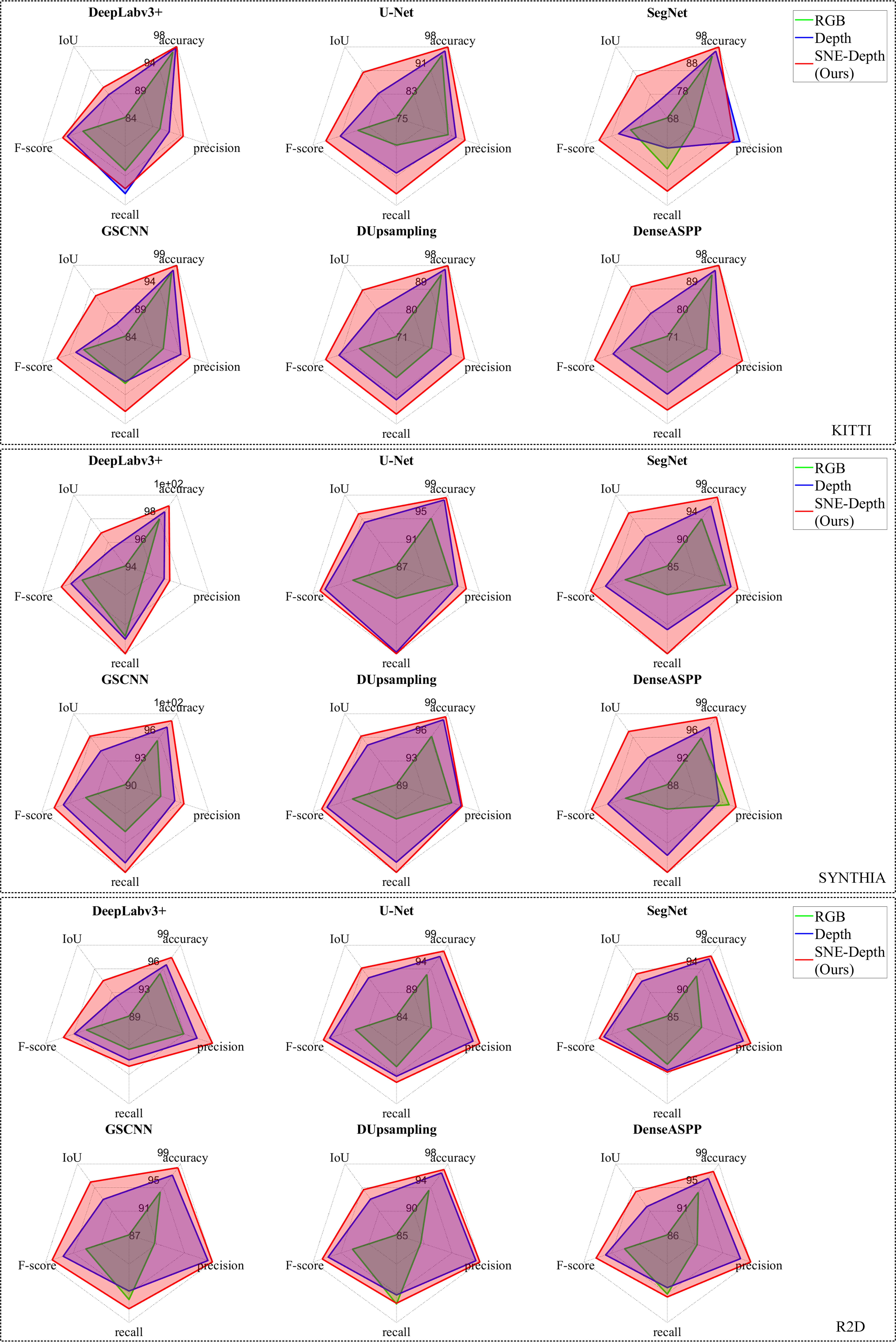}
    \caption{Performance comparison ($\%$) among DeepLabv3+ \cite{chen2018encoder}, U-Net \cite{ronneberger2015u}, SegNet \cite{badrinarayanan2017segnet}, GSCNN \cite{takikawa2019gated}, DUpsampling \cite{tian2019decoders} and DenseASPP \cite{yang2018denseaspp} with and without our SNE embedded, where \protect\greenline\ RGB, \protect\blueline\ Depth, and \protect\redline\ SNE-Depth (Ours). }
    \label{fig.radars}
\end{figure*}

\begin{figure*}[!t]
    \centering
    \includegraphics[width=0.82\textwidth]{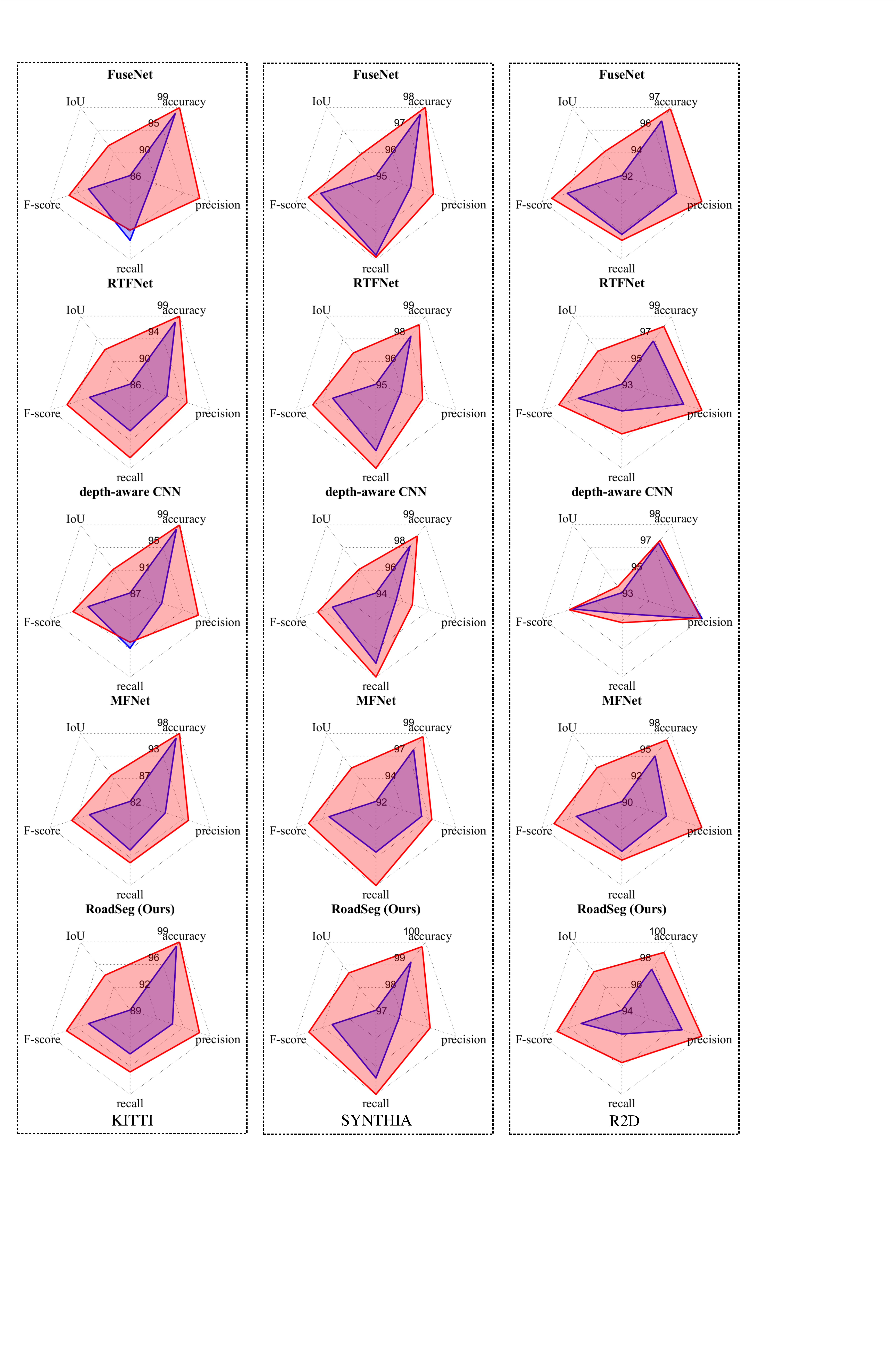}
    \caption{Performance comparison ($\%$) among FuseNet \cite{hazirbas2016fusenet}, RTFNet \cite{sun2019rtfnet}, depth-aware CNN \cite{wang2018depth}, MFNet \cite{ha2017mfnet} and our RoadSeg with and without our SNE embedded, where \protect\blueline\ RGBD and \protect\redline\ SNE-RGBD (Ours).}
    \label{fig.two_stream}
\end{figure*}

\subsection{Performance Evaluation of Our SNE}
\label{sec.performance_evaluation_of_sne}

We simply set  $g_x=\frac{1}{Z(x-1,y)}-\frac{1}{Z(x+1,y)}$ and $g_y=\frac{1}{Z(x,y-1)}-\frac{1}{Z(x,y+1)}$ to evaluate the accuracy of our proposed SNE. In addition, we also compare it with two well-known surface normal estimation approaches: SRI \cite{badino2011fast} and LINE-MOD \cite{hinterstoisser2011gradient}. The qualitative and quantitative comparisons are shown in Fig. \ref{fig.diode}. It can be observed that our proposed SNE outperforms SRI and LINE-MOD for both indoor and outdoor scenarios.

\subsection{Performance Evaluation of Our SNE-RoadSeg}
\label{sec.performance_evaluation_of_roadseg}

\begin{table*}[!t]
    \centering
    \caption{The KITTI road benchmark results, where the best results are in bold type. Please note that we only compare our method with published works.}
    \label{tab.kittiroad}
    \setlength\arrayrulewidth{0.7pt}
    \begin{tabular}{L{10em}|C{5em}|C{5em}|C{5em}|C{5em}|C{5em}}
    \hline
    Method &
    MaxF ($\%$) &
    AP ($\%$)&
    PRE ($\%$) &
    REC ($\%$) &
    Rank \\
    \hline

    RBNet \cite{chen2017rbnet}
    & 93.21 & 89.18 & 92.81 & 93.60 & 21
    \\

    TVFNet \cite{gu2019two}
    & 95.34 & 90.26 & 95.73 & 94.94 & 16
    \\

    LC-CRF \cite{gu2019road}
    & 95.68 & 88.34 & 93.62 & \textbf{97.83} & 13
    \\

    LidCamNet \cite{caltagirone2019lidar}
    & 96.03 & 93.93 & 96.23 & 95.83 & 7
    \\

    RBANet \cite{sun2019reverse}
    & 96.30 & 89.72 & 95.14 & 97.50 & 6
    \\
    \hline

    SNE-RoadSeg (Ours)
    & \textbf{96.75} & \textbf{94.07} & \textbf{96.90} & 96.61 & \textbf{2}
    \\
    \hline

    \end{tabular}
\end{table*}

\begin{figure*}[t]
    \centering
    \includegraphics[width=\textwidth]{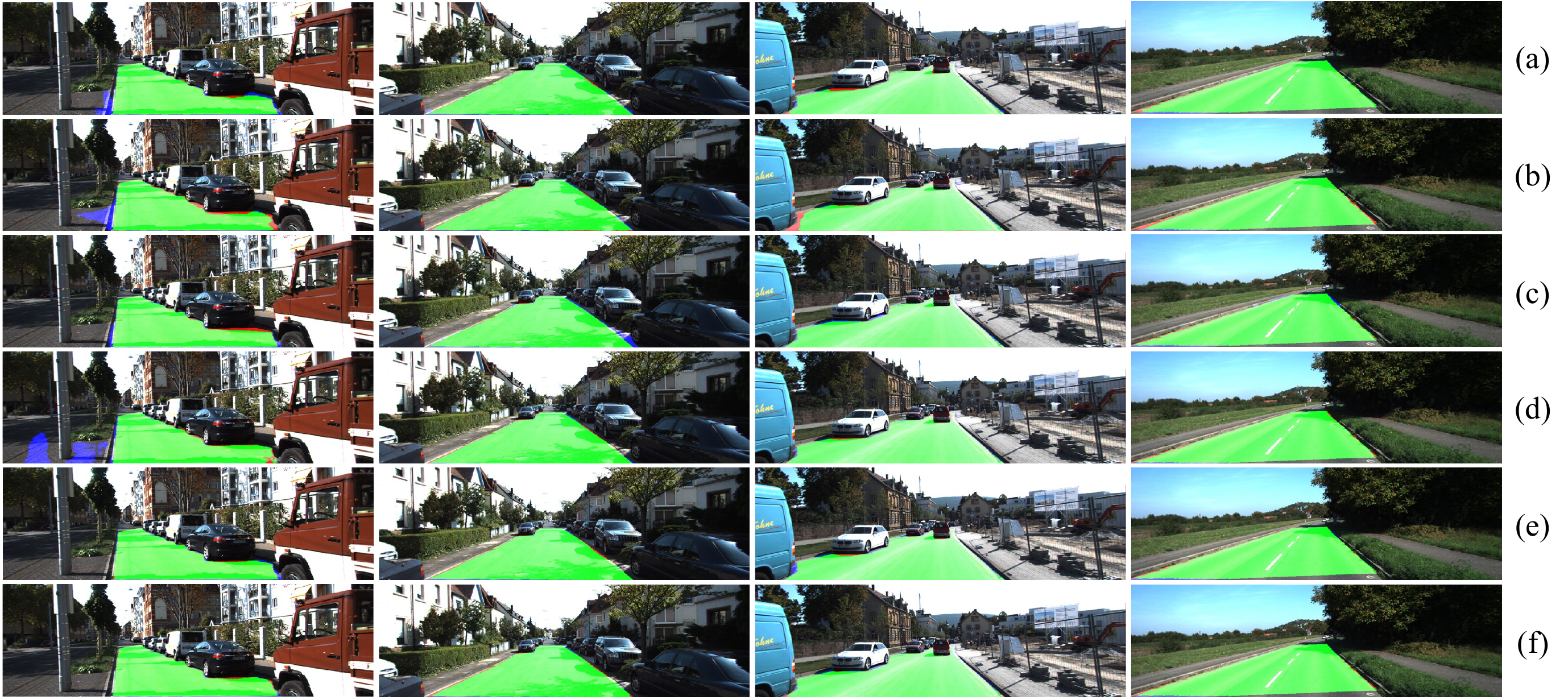}
    \caption{Examples on the KITTI road benchmark, where rows (a)--(f) show the freespace detection results obtained by RBNet \cite{chen2017rbnet}, TVFNet \cite{gu2019two}, LC-CRF \cite{gu2019road}, LidCamNet \cite{caltagirone2019lidar}, RBANet \cite{sun2019reverse} and our proposed SNE-RoadSeg, respectively. The true positive, false negative and false positive pixels are shown in green, red and blue, respectively.}
    \label{fig.kittiroad}
\end{figure*}

In this subsection, we evaluate the performance of our proposed SNE-RoadSeg-152 (abbreviated as SNE-RoadSeg) both qualitatively and quantitatively. Examples of the experimental results on the SYNTHIA road dataset \cite{HernandezBMVC17} and our R2D road dataset are shown in Fig. \ref{fig.synthiapluscarla}. We can clearly observe that the CNNs with RGB images as inputs suffer greatly from poor illumination conditions. Moreover, the CNNs with our SNE embedded generally perform better than they do without our SNE embedded. The corresponding quantitative comparisons are given in Fig \ref{fig.radars} and Fig. \ref{fig.two_stream}. Readers can see that the IoU increases by approximately 2-12\% for single-modal CNNs and by about 1-7\% for data-fusion CNNs, while the F-score increases by around 1-7\% for single-modal CNNs and by about 1-4\% for data-fusion CNNs. We demonstrate that our proposed SNE can make the road areas become highly distinguishable, and thus, it will benefit all state-of-the-art CNNs for freespace detection.

Furthermore, from Fig \ref{fig.radars} and Fig. \ref{fig.two_stream}, we can observe that RoadSeg itself outperforms all other CNNs. We demonstrate that the densely-connected skip connections utilized in our proposed RoadSeg can help achieve flexible feature fusion and smooth the gradient flow to generate accurate freespace detection results. Also, RoadSeg with our SNE embedded performs better than all other CNNs with our SNE embedded. An increase of approximately 1.4-14.7\% is witnessed on the IoU, while the F-score increases by about 0.7-8.8\%.

In addition, we compare our proposed method with five state-of-the-art CNNs published on the KITTI road benchmark \cite{Fritsch2013ITSC}. Examples of the experimental results are shown in Fig. \ref{fig.kittiroad}. The quantitative comparisons are given in Table \ref{tab.kittiroad}, which shows that our proposed SNE-RoadSeg achieves the highest MaxF (maximum F-score), AP (average precision) and PRE (precision), while LC-CRF \cite{gu2019road} achieves the best REC (recall). Our freespace detection method is the {second best} on the KITTI road benchmark \cite{Fritsch2013ITSC}.

\begin{figure*}[t]
    \centering
    \includegraphics[width=\textwidth]{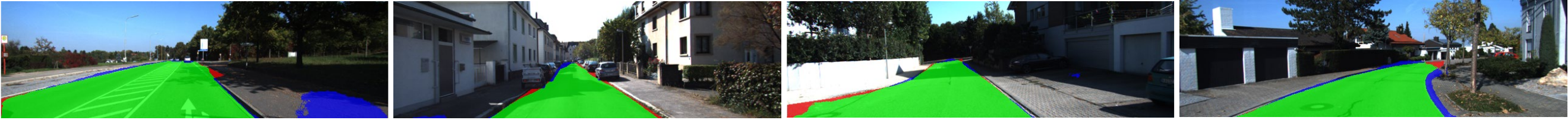}
    \caption{Unsatisfactory results obtained using the KITTI road dataset. The true positive, false negative and false positive pixels are shown in green, red and blue, respectively.}
    \label{fig.failure}
\end{figure*}

Fig. \ref{fig.failure} presents several unsatisfactory results of our SNE-RoadSeg on the KITTI road dataset \cite{Fritsch2013ITSC}. Since the 3D points on freespace and sidewalks possess very similar surface normals, our proposed approach can sometimes mistakenly recognize part of sidewalks as freespace, especially when the textures of the road and sidewalks are similar. We believe this can be improved by leveraging surface normal gradient features, as there usually exists a clear boundary between freespace and sidewalks (due to their differences in height).

\subsection{Ablation Study}
\label{sec.ablation_study}

In this subsection, we conduct ablation studies on our R2D road dataset to validate the effectiveness of the architecture for our RoadSeg. The performances of different architectures are provided in Table \ref{tab.ablation}.

Firstly, we replace the backbone of RoadSeg with different ResNet architectures. The quantitative results are given in Table \ref{tab.ablation}. The superior performance of our choice is as expected, because ResNet-152 has also presented the best image classification performance among the five ResNet architectures \cite{he2016deep}.

Then, we remove one encoder from RoadSeg to evaluate its performance on single-modal vision data. We conduct five experiments: a) training with RGB images, denoted as \textbf{RGB}; b) training with depth images, denoted as \textbf{Depth}; c) training with depth images, denoted as \textbf{SNE-Depth}; d) training with four-channel RGB-D vision data, denoted as \textbf{RGBD-C}; and e) training with four-channel RGB-D vision data, denoted as \textbf{SNE-RGBD-C}. From Table \ref{tab.ablation}, we can observe that our choice outperforms the single-modal architecture with respect to different modalities of training data, proving that the data fusion via a two-encoder architecture can benefit the freespace detection. It should be noted that although the single-modal architectures cannot provide competitive results, our proposed SNE still benefits them for better freespace detection performance.

\begin{table*}[t]
    \centering
    \caption{Performance comparison ($\%$) among different architectures and setups on our R2D road dataset. The best results are shown in bold font.}
    \label{tab.ablation}
    \setlength\arrayrulewidth{0.7pt}
    \begin{tabular}{L{9em}|L{6.5em}|C{4em}|C{4em}|C{4em}|C{4em}|C{3.5em}}
        \hline
        Architecture &
        Setup &
        Accuracy &
        Precision &
        Recall &
        F-Score &
        IoU \\
        \hline
        
        RoadSeg-18
        & \multirow{4}{*}{SNE-RGBD} & 93.6 & 93.5 & 91.3 & 92.4 & 85.9
        \\
        
        RoadSeg-34
        & & 95.5 & 96.3 & 93.0 & 94.6 & 89.8
        \\
        
        RoadSeg-50
        & & 96.8 & 97.5 & 95.2 & 96.3 & 92.9
        \\
        
        RoadSeg-101
        & & 98.0 & 98.2 & 97.1 & 97.6 & 95.4
        \\
        \hline
        
        \multirow{5}{*}{RoadSeg-152}
        & RGB & 94.0 & 91.9 & 93.8 & 92.8 & 86.6
        \\
        
        & Depth & 96.7 & 97.6 & 94.6 & 96.1 & 92.4
        \\
        
        & SNE-Depth & 97.6 & 98.9 & 95.5 & 97.2 & 94.5
        \\
        
        & RGBD-C & 95.1 & 92.8 & 95.6 & 94.2 & 89.0
        \\
        
        & SNE-RGBD-C & 97.0 & 97.5 & 95.3 & 96.4 & 93.0
        \\
        \hline
        
        RoadSeg-152-NSCs
        & \multirow{2}{*}{SNE-RGBD} & 97.9 & 98.6 & 96.5 & 97.5 & 95.2
        \\
        
        RoadSeg-152-SSCs
        & & 98.2 & 99.0 & 96.8 & 97.9 & 95.9
        \\
        \hline
        
        RoadSeg-152 (Ours)
        & SNE-RGBD & \textbf{98.6} & \textbf{99.1} & \textbf{97.6} & \textbf{98.3} & \textbf{96.7}
        \\
        \hline
    \end{tabular}
\end{table*}

To further validate the effectiveness of our choice, we replace the densely-connected skip connections in the decoder with two different architectures:
a) no skip connections (NSCs), which totally removes the skip connections; b) sparse skip connections (SSCs), which employs the skip connections only at the same-scale feature maps of the encoder and decoder (like U-Net). Table \ref{tab.ablation} verifies the superiority of the densely-connected skip connections, which helps to achieve flexible feature fusion and to smooth the gradient flow to generate accurate freespace detection results, as analyzed in Section \ref{sec.performance_evaluation_of_roadseg}.

\section{Conclusion}
\label{sec.conclusion}
The main contributions of this paper include: a) a module named SNE, capable of inferring surface normal information from depth/disparity images with both high precision and efficiency; b) a data-fusion CNN architecture named RoadSeg, capable of fusing both RGB and surface normal information for accurate freespace detection; and c) a publicly available synthetic dataset for semantic driving scene segmentation. To demonstrate the feasibility and effectiveness of the proposed SNE module, we embedded it into ten state-of-the-art CNNs and evaluated their performances for freespace detection. The experimental results illustrated that our introduced SNE can benefit all these CNNs for freespace detection. Furthermore, our proposed data-fusion CNN architecture RoadSeg is most compatible with our proposed SNE, and it outperforms all other CNNs when detecting drivable road regions.

\section*{Acknowledgements}
This work was supported by the National Natural Science Foundation of China, under grant No. U1713211, and the Research Grant Council of Hong Kong SAR Government, China, under Project No. 11210017, awarded to Prof. Ming Liu.

\clearpage
\bibliographystyle{splncs04}

\begin{thebibliography}{10}
\providecommand{\url}[1]{\texttt{#1}}
\providecommand{\urlprefix}{URL }
\providecommand{\doi}[1]{https://doi.org/#1}

\bibitem{alvarez2012road}
Alvarez, J.M., Gevers, T., LeCun, Y., Lopez, A.M.: Road scene segmentation from
  a single image. In: European Conference on Computer Vision. pp. 376--389.
  Springer (2012)

\bibitem{badino2011fast}
Badino, H., Huber, D., Park, Y., Kanade, T.: Fast and accurate computation of
  surface normals from range images. In: 2011 IEEE International Conference on
  Robotics and Automation. pp. 3084--3091. IEEE (2011)

\bibitem{badrinarayanan2017segnet}
Badrinarayanan, V., Kendall, A., Cipolla, R.: Segnet: A deep convolutional
  encoder-decoder architecture for image segmentation. IEEE transactions on
  pattern analysis and machine intelligence  \textbf{39}(12),  2481--2495
  (2017)

\bibitem{cai2020prob}
Cai, P., Wang, S., Sun, Y., Liu, M.: Probabilistic end-to-end vehicle
  navigation in complex dynamic environments with multimodal sensor fusion.
  IEEE Robotics and Automation Letters  \textbf{5},  4218--4224 (2020)

\bibitem{caltagirone2019lidar}
Caltagirone, L., Bellone, M., Svensson, L., Wahde, M.: Lidar--camera fusion for
  road detection using fully convolutional neural networks. Robotics and
  Autonomous Systems  \textbf{111},  125--131 (2019)

\bibitem{chen2014semantic}
Chen, L.C., Papandreou, G., Kokkinos, I., Murphy, K., Yuille, A.L.: Semantic
  image segmentation with deep convolutional nets and fully connected crfs.
  CoRR  \textbf{abs/1412.7062} (2014)

\bibitem{chen2017deeplab}
Chen, L.C., Papandreou, G., Kokkinos, I., Murphy, K., Yuille, A.L.: Deeplab:
  Semantic image segmentation with deep convolutional nets, atrous convolution,
  and fully connected crfs. IEEE transactions on pattern analysis and machine
  intelligence  \textbf{40}(4),  834--848 (2017)

\bibitem{chen2017rethinking}
Chen, L.C., Papandreou, G., Schroff, F., Adam, H.: Rethinking atrous
  convolution for semantic image segmentation. arXiv preprint arXiv:1706.05587
  (2017)

\bibitem{chen2018encoder}
Chen, L.C., Zhu, Y., Papandreou, G., Schroff, F., Adam, H.: Encoder-decoder
  with atrous separable convolution for semantic image segmentation. In:
  Proceedings of the European conference on computer vision (ECCV). pp.
  801--818 (2018)

\bibitem{chen2017rbnet}
Chen, Z., Chen, Z.: Rbnet: A deep neural network for unified road and road
  boundary detection. In: International Conference on Neural Information
  Processing. pp. 677--687. Springer (2017)

\bibitem{dosovitskiy2017carla}
Dosovitskiy, A., Ros, G., Codevilla, F., Lopez, A., Koltun, V.: {CARLA}: {An}
  open urban driving simulator. In: Levine, S., Vanhoucke, V., Goldberg, K.
  (eds.) Proceedings of the 1st Annual Conference on Robot Learning.
  Proceedings of Machine Learning Research, vol.~78, pp. 1--16. PMLR (13--15
  Nov 2017)

\bibitem{fan2019uav}
{Fan}, R., {Jiao}, J., {Pan}, J., {Huang}, H., {Shen}, S., {Liu}, M.: Real-time
  dense stereo embedded in a {UAV} for road inspection. In: Proc. IEEE/CVF
  Conf. Computer Vision and Pattern Recognition Workshops (CVPRW). pp. 535--543
  (2019)

\bibitem{fan2019pothole}
Fan, R., Ozgunalp, U., Hosking, B., Liu, M., Pitas, I.: Pothole detection based
  on disparity transformation and road surface modeling. IEEE Transactions on
  Image Processing  \textbf{29},  897--908 (2019)

\bibitem{fan2020three}
Fan, R., Wang, H., Xue, B., Huang, H., Wang, Y., Liu, M., Pitas, I.:
  Three-filters-to-normal: An accurate and ultrafast surface normal estimator.
  arXiv preprint arXiv:2005.08165  (2020), under peer review

\bibitem{Fritsch2013ITSC}
Fritsch, J., Kuehnl, T., Geiger, A.: A new performance measure and evaluation
  benchmark for road detection algorithms. In: International Conference on
  Intelligent Transportation Systems (ITSC) (2013)

\bibitem{gu2019road}
Gu, S., Zhang, Y., Tang, J., Yang, J., Kong, H.: Road detection through crf
  based lidar-camera fusion. In: 2019 International Conference on Robotics and
  Automation (ICRA). pp. 3832--3838. IEEE (2019)

\bibitem{gu2019two}
Gu, S., Zhang, Y., Yang, J., Alvarez, J.M., Kong, H.: Two-view fusion based
  convolutional neural network for urban road detection. In: 2019 IEEE/RSJ
  International Conference on Intelligent Robots and Systems (IROS). pp.
  6144--6149. IEEE (2019)

\bibitem{ha2017mfnet}
Ha, Q., Watanabe, K., Karasawa, T., Ushiku, Y., Harada, T.: Mfnet: Towards
  real-time semantic segmentation for autonomous vehicles with multi-spectral
  scenes. In: 2017 IEEE/RSJ International Conference on Intelligent Robots and
  Systems (IROS). pp. 5108--5115. IEEE (2017)

\bibitem{hazirbas2016fusenet}
Hazirbas, C., Ma, L., Domokos, C., Cremers, D.: Fusenet: Incorporating depth
  into semantic segmentation via fusion-based cnn architecture. In: Asian
  conference on computer vision. pp. 213--228. Springer (2016)

\bibitem{he2016deep}
He, K., Zhang, X., Ren, S., Sun, J.: Deep residual learning for image
  recognition. In: Proceedings of the IEEE conference on computer vision and
  pattern recognition. pp. 770--778 (2016)

\bibitem{HernandezBMVC17}
Hernandez-Juarez, D., Schneider, L., Espinosa, A., Vazquez, D., Lopez, A.M.,
  Franke, U., Pollefeys, M., Moure, J.C.: Slanted stixels: Representing san
  francisco’s steepest streets. In: British Machine Vision Conference (BMVC),
  2017 (2017)

\bibitem{hinterstoisser2011gradient}
Hinterstoisser, S., Cagniart, C., Ilic, S., Sturm, P., Navab, N., Fua, P.,
  Lepetit, V.: Gradient response maps for real-time detection of textureless
  objects. IEEE transactions on pattern analysis and machine intelligence
  \textbf{34}(5),  876--888 (2011)

\bibitem{huang2017densely}
Huang, G., Liu, Z., Van Der~Maaten, L., Weinberger, K.Q.: Densely connected
  convolutional networks. In: Proceedings of the IEEE conference on computer
  vision and pattern recognition. pp. 4700--4708 (2017)

\bibitem{long2015fully}
Long, J., Shelhamer, E., Darrell, T.: Fully convolutional networks for semantic
  segmentation. In: Proceedings of the IEEE conference on computer vision and
  pattern recognition. pp. 3431--3440 (2015)

\bibitem{lu2019monocular}
Lu, C., van~de Molengraft, M.J.G., Dubbelman, G.: Monocular semantic occupancy
  grid mapping with convolutional variational encoder--decoder networks. IEEE
  Robotics and Automation Letters  \textbf{4}(2),  445--452 (2019)

\bibitem{ronneberger2015u}
Ronneberger, O., Fischer, P., Brox, T.: U-net: Convolutional networks for
  biomedical image segmentation. In: International Conference on Medical image
  computing and computer-assisted intervention. pp. 234--241. Springer (2015)

\bibitem{sless2019road}
Sless, L., El~Shlomo, B., Cohen, G., Oron, S.: Road scene understanding by
  occupancy grid learning from sparse radar clusters using semantic
  segmentation. In: Proceedings of the IEEE International Conference on
  Computer Vision Workshops. pp.~0--0 (2019)

\bibitem{sun2019reverse}
Sun, J.Y., Kim, S.W., Lee, S.W., Kim, Y.W., Ko, S.J.: Reverse and boundary
  attention network for road segmentation. In: Proceedings of the IEEE
  International Conference on Computer Vision Workshops. pp.~0--0 (2019)

\bibitem{sun2019rtfnet}
Sun, Y., Zuo, W., Liu, M.: Rtfnet: Rgb-thermal fusion network for semantic
  segmentation of urban scenes. IEEE Robotics and Automation Letters
  \textbf{4}(3),  2576--2583 (2019)

\bibitem{takikawa2019gated}
Takikawa, T., Acuna, D., Jampani, V., Fidler, S.: Gated-scnn: Gated shape cnns
  for semantic segmentation. In: Proceedings of the IEEE International
  Conference on Computer Vision. pp. 5229--5238 (2019)

\bibitem{thoma2019mapping}
Thoma, J., Paudel, D.P., Chhatkuli, A., Probst, T., Gool, L.V.: Mapping,
  localization and path planning for image-based navigation using visual
  features and map. In: Proceedings of the IEEE Conference on Computer Vision
  and Pattern Recognition. pp. 7383--7391 (2019)

\bibitem{tian2019decoders}
Tian, Z., He, T., Shen, C., Yan, Y.: Decoders matter for semantic segmentation:
  Data-dependent decoding enables flexible feature aggregation. In: Proceedings
  of the IEEE Conference on Computer Vision and Pattern Recognition. pp.
  3126--3135 (2019)

\bibitem{vasiljevic2019diode}
Vasiljevic, I., Kolkin, N., Zhang, S., Luo, R., Wang, H., Dai, F.Z., Daniele,
  A.F., Mostajabi, M., Basart, S., Walter, M.R., et~al.: Diode: A dense indoor
  and outdoor depth dataset. arXiv preprint arXiv:1908.00463  (2019)

\bibitem{wang2020applying}
Wang, H., Fan, R., Sun, Y., Liu, M.: Applying surface normal information in
  drivable area and road anomaly detection for ground mobile robots. In: 2020
  IEEE/RSJ International Conference on Intelligent Robots and Systems (IROS)
  (2020), to be published

\bibitem{wang2018depth}
Wang, W., Neumann, U.: Depth-aware cnn for rgb-d segmentation. In: Proceedings
  of the European Conference on Computer Vision (ECCV). pp. 135--150 (2018)

\bibitem{wedel2009b}
Wedel, A., Badino, H., Rabe, C., Loose, H., Franke, U., Cremers, D.: B-spline
  modeling of road surfaces with an application to free-space estimation. IEEE
  transactions on Intelligent transportation systems  \textbf{10}(4),  572--583
  (2009)

\bibitem{yang2018denseaspp}
Yang, M., Yu, K., Zhang, C., Li, Z., Yang, K.: Denseaspp for semantic
  segmentation in street scenes. In: Proceedings of the IEEE Conference on
  Computer Vision and Pattern Recognition. pp. 3684--3692 (2018)

\end{thebibliography}

\end{document}